# A unified framework for global and local interpretability using adaptive derivative-ordered random explanation

Lemen Chao[a, b] Ming Lei[a] Anran Fang[a,*]

[a] *School of Information Resource Management, Renmin University of China, Beijing, 100872, People's Republic of China.*

[b] *Key Laboratory of Data Engineering and Knowledge Engineering, Beijing, 100872, People's Republic of China.*

[*]Corresponding author. E-mail address: fanganran97@126.com (Anran Fang)

**Abstract:** The interpretability of complex machine learning models is of paramount importance, especially in real-world high-stakes domains such as healthcare and finance. However, existing post-hoc interpretability methods suffer from inherent limitations: fragmented analytical processes, inadequate capacity to model nonlinear feature interactions, computational inefficiencies, and over-reliance on specific model architectures. To address these challenges, this paper provides a novel method—Adaptive Derivative-Ordered Random Explanation (ADORE)—that leverages first- and second-order derivatives to accommodate nonlinear model complexities, while enabling effective capture of feature-sample interactions within a unified analytical framework. ADORE integrates global feature importance with local sample contributions, precisely quantifying feature impact by capturing both magnitude and direction, and identifying critical samples influencing model decisions. Furthermore, it achieves computational efficiency through randomized singular value decomposition (SVD) and dynamic sparsity detection, making it scalable to large, high-dimensional datasets. Experiments across three data modalities—tabular, text, and image—demonstrate that ADORE outperforms existing methods such as LIME and SHAP in handling complex interactions and computational efficiency, while providing detailed and reliable explanations. To facilitate adoption and reproducibility, ADORE have been released as an open-source Python package[1], hosted on GitHub, enabling researchers and practitioners to readily adapt and apply our approach to their specific tasks, models, and datasets.

[1] https://pypi.org/project/adore/0.1.2/

## 1. Introduction

The advent of complex machine learning models, including deep neural networks and ensemble methods, has significantly advanced predictive performance across various domains. Despite their strong performance in real-world applications, these models often rely heavily on large datasets and intricate hyperparameter tuning. Their feature learning and decision-making processes are difficult to interpret, lacking clear mathematical underpinnings and relying predominantly on trial-and-error optimization strategies[19][2][25]. This "black-box" nature undermines the trustworthiness of such models in practical scenarios and raises concerns about their lack of interpretability. In high-stakes decision-making areas such as healthcare, finance, and cybersecurity, interpretability is not just desirable—it is imperative[26]. It fosters trust, ensures compliance with regulatory standards, and enables practitioners to understand and mitigate potential biases within the models[18].

Interpretability techniques have been continuously evolving, with current approaches broadly categorized into four types: active interpretability, passive interpretability, supplementary interpretability, and integrative interpretability[31]. Among these, passive interpretability, exemplified by post-hoc interpretability methods, stands out due to its broader applicability and greater flexibility. These methods aim to extract logical rules or comprehensible patterns by analyzing a model's structure, weights, and outputs. However, mainstream post-hoc interpretability methods still face fundamental challenges, which limit both their effectiveness in practical applications and their broader adoption in real-world scenarios.

First, existing methods often isolate individual features or samples during analysis. This fragmented approach overlooks the dynamic interactions between features and across data instances. By failing to provide a global perspective on model behavior, these methods tend to produce oversimplified interpretations that can sometimes be misleading[29]. Second, these methods exhibit significant limitations in capturing complex nonlinear relationships between features, which are critical for accurate predictions and reliable decision-making in models like deep neural networks and ensemble methods. Third, computational inefficiency remains a major obstacle. Methods such as SHAP, while theoretically robust, suffer from high computational complexity when applied to high-dimensional data or large datasets, thereby constraining their practicality in real-world scenarios[21]. Finally, model dependency restricts the generalizability of certain interpretability methods[6]. For instance, methods like Tree SHAP are specifically tailored for tree-based models and may not be applicable to neural networks. These challenges underscore the urgent need for more integrated, computationally efficient, and universally applicable interpretability solutions.

To address these challenges, this study proposes **ADORE** (Adaptive Derivative Order Randomized

Explanation), a novel algorithm that provides a unified explainability framework integrating both global feature importance and local sample contributions. The main contributions of this work are:

- **Unified Interpretations**: ADORE constructs a derivative matrix that simultaneously captures global feature importance and local sample contributions, overcoming the limitations of fragmented analyses.
- **Adaptive Derivative Order Selection**: The algorithm dynamically switches between first-order and second-order derivatives based on model complexity, ensuring accurate interpretations for both linear and highly nonlinear models.
- **Efficient Computation via Randomized SVD**: By employing Randomized Singular Value Decomposition and dynamic sparsity detection, ADORE significantly reduces computational complexity and memory consumption, enabling scalability to large datasets.
- **Model-Agnostic Applicability**: ADORE is applicable to a wide range of black-box models without reliance on internal structures, enhancing its versatility.

The remainder of the paper is structured as follows: Section 2 reviews existing approaches to model interpretability, highlighting their strengths, limitations, and situating the ADORE algorithm within the current state-of-the-art. Section 3 presents the ADORE algorithm, outlining its conceptual framework and mathematical underpinnings. Section 4 details the experimental setup, including three modality-based prediction tasks, baseline methods, evaluation metrics, parameters, and environment. Section 5 presents the results, comparing ADORE with existing methods to assess both consistency and differences. Section 6 analyzes computational complexity, discusses theoretical and practical implications, outlines limitations, and suggests future research directions. Finally, Section 7 concludes the study.

## 2. Related Work

Currently, the classification approaches in interpretability research exhibit significant diversity. Scholars have categorized interpretability methods across various dimensions, often based on model transparency and functionality. These dimensions include active versus passive methods, local versus global interpretability, application stages, scope of interpretability, and model dependency, among others[4][7][17][33][5]. According to a recent survey[31], interpretability methods can be broadly divided into four categories: active, passive, supplementary, and integrated explanations.

Active interpretability involves modifying or optimizing the model architecture before training to improve transparency, while passive interpretability leverages inherently transparent models or post-hoc analysis without structural changes. Post-hoc methods, such as hidden layer analysis and class activation mapping, are highly versatile as they can be applied directly to pre-trained models. Since our proposed algorithm belongs to this category, the following section provides a systematic review and critical analysis of post-hoc approaches from both local and global perspectives.

### 2.1 Local post-hoc explainability methods

Local post-hoc interpretability methods aim to explain individual predictions by approximating the model's behavior in the vicinity of a specific instance. One prominent approach is LIME (Local Interpretable Model-agnostic Explanations), proposed by[23]. LIME operates by constructing an interpretable surrogate model, typically a linear model, to locally approximate the behavior of the black-box model around a given prediction. It achieves this by perturbing input features and observing the resulting changes in output, thereby estimating the contribution of each feature to the prediction. However, LIME's reliance on linear approximation may limit its ability to capture complex nonlinear interactions that are often present in more sophisticated models.

Several researchers have optimized and extended LIME to improve its stability, local fidelity, and applicability. For instance, [27] introduced a local interpretability method based on autoencoders, incorporating autoencoders into the design of the weighting function. This significantly enhanced LIME's stability and local fidelity when explaining predictions from deep learning models. [22] developed a Case–Based Reasoning–LIME (CBR-LIME) tailored for image classifier interpretability, addressing challenges related to parameter configuration in the LIME algorithm. Additionally, [14] proposed GraphLIME, a specialized local interpretability method designed to account for the structural characteristics of graph neural networks (GNNs).

Another influential post-hoc interpretability method is SHAP (Shapley Additive Explanations), developed by [20]. SHAP leverages concepts from cooperative game theory to compute the marginal contribution of each feature to the model's output, represented as Shapley values. The method theoretically offers a robust attribution mechanism and provides a detailed analysis of feature importance. However, while SHAP excels in interpretability, its computational complexity is substantial, particularly when applied to large datasets or models with high-dimensional feature spaces, resulting in significant computational costs.

Building on SHAP, [9] integrated the eXtreme Gradient Boosting (XGBoost) algorithm with SHAP to construct interpretable machine learning models for structural engineering applications. This approach enhanced both prediction accuracy and model transparency. [10] developed DeepSHAP, an extension designed to explain neural retrieval models used in information retrieval tasks. DeepSHAP builds upon the DeepLIFT algorithm by estimating Shapley values for input features, quantifying their contributions to the difference between the prediction result and the average prediction value. Notably, SHAP's computational time increases exponentially with the number of features, and the order of feature combinations can influence the calculation of Shapley values, potentially impacting the accuracy of the model's explanations.

Additionally, [24] introduced the concept of anchors, which provide high-precision if-then rules to "anchor" predictions. These rules exhibit local reliability and hold with high probability, offering intuitive insights into the model's behavior for specific instances. However, while anchor models excel at delivering clear and precise explanations, they may fall short in capturing the model's global behavior or the intricate interactions between features and samples.

### 2.2 Global post-hoc explainability methods

Global interpretation methods aim to provide insights into the overall behavior of a model, offering a broader understanding beyond individual predictions. One common approach is Feature Importance Rankings, which aggregate measures of feature influence across the entire dataset. Metrics such as mean decrease in impurity for tree-based models or permutation importance, as introduced by [3], are often used. While these rankings are valuable for identifying influential features, they may overlook complex interactions between features and do not offer explanations specific to individual instances. To address the need for more nuanced insights, Partial Dependence Plots (PDPs), proposed by [11], are employed to visualize the marginal effect of one or two features on the predicted outcome while holding other features constant. PDPs are useful for understanding the average effect of features, but they can be misleading when features are correlated or when there are intricate interactions within the model. In response to some of the limitations of PDPs, [1] introduced Accumulated Local Effects (ALE) Plots as an alternative that takes into account feature interactions and mitigates bias caused by correlated features. Although ALE plots provide a more accurate depiction of global patterns, they still emphasize overall trends and may not capture local variations that are critical for certain applications. Together, these global methods highlight the ongoing challenge of balancing comprehensive model interpretation with the need to account for both global and local patterns, underscoring the complexity of extracting meaningful insights from complex models.

Recognizing the limitations of strictly local or global methods, some approaches attempt to unify these perspectives. Gradient-Based Methods, such as Integrated Gradients[30] and DeepLIFT[28], provide attribution scores by analyzing gradients of the output with respect to input features. These methods can capture both local and global interpretations to some extent but may struggle with models exhibiting highly nonlinear behaviors or may not offer precise quantification of feature contributions.

One major issue with existing methods is their fragmented approach to analysis, as these methods often focus on either local or global explanations while neglecting the critical interdependencies between features and samples.

This compartmentalized perspective leads to incomplete understanding, failing to capture the context-dependent feature interactions necessary for accurate interpretation. Moreover, local interpretability methods, such as LIME, struggle to adequately model nonlinear interactions. These methods rely on linear approximations, which are inherently limited in representing the complex nonlinear relationships characteristic of models like deep neural networks and ensemble methods. This limitation can result in misleading explanations, particularly in scenarios where feature interactions play a significant role in influencing predictions.

Another key limitation is computational inefficiency. High-complexity methods like SHAP require evaluating all possible feature combinations, leading to prohibitive computational costs when dealing with high-dimensional data or large-scale datasets. Additionally, the reliance of interpretability methods on specific model types presents further challenges. For instance, model-specific approaches such as GraphLIME and TreeSHAP are tailored for particular models, which restricts their applicability across diverse model types.

In summary, these limitations underscore the need for more generalizable and comprehensive post-hoc interpretability methods that can address fragmented analysis, capture nonlinear feature interactions, ensure computational efficiency, and maintain broad applicability across various model architectures.

## 3. The ADORE Algorithm

ADORE presents a unified framework that integrates global feature importance with local sample contributions. First, an adaptive perturbation strategy is applied to the sample feature matrix to generate a perturbation matrix. Then, a derivative matrix is constructed to capture the complex relationships between features and predictions across samples, accurately representing both linear and nonlinear interactions. Next, random singular value decomposition (random SVD) is employed to compute and store the derivative matrix. Finally, contribution values are calculated at different levels of granularity. Fig. 1 shows the basic logical framework of the ADORE algorithm.

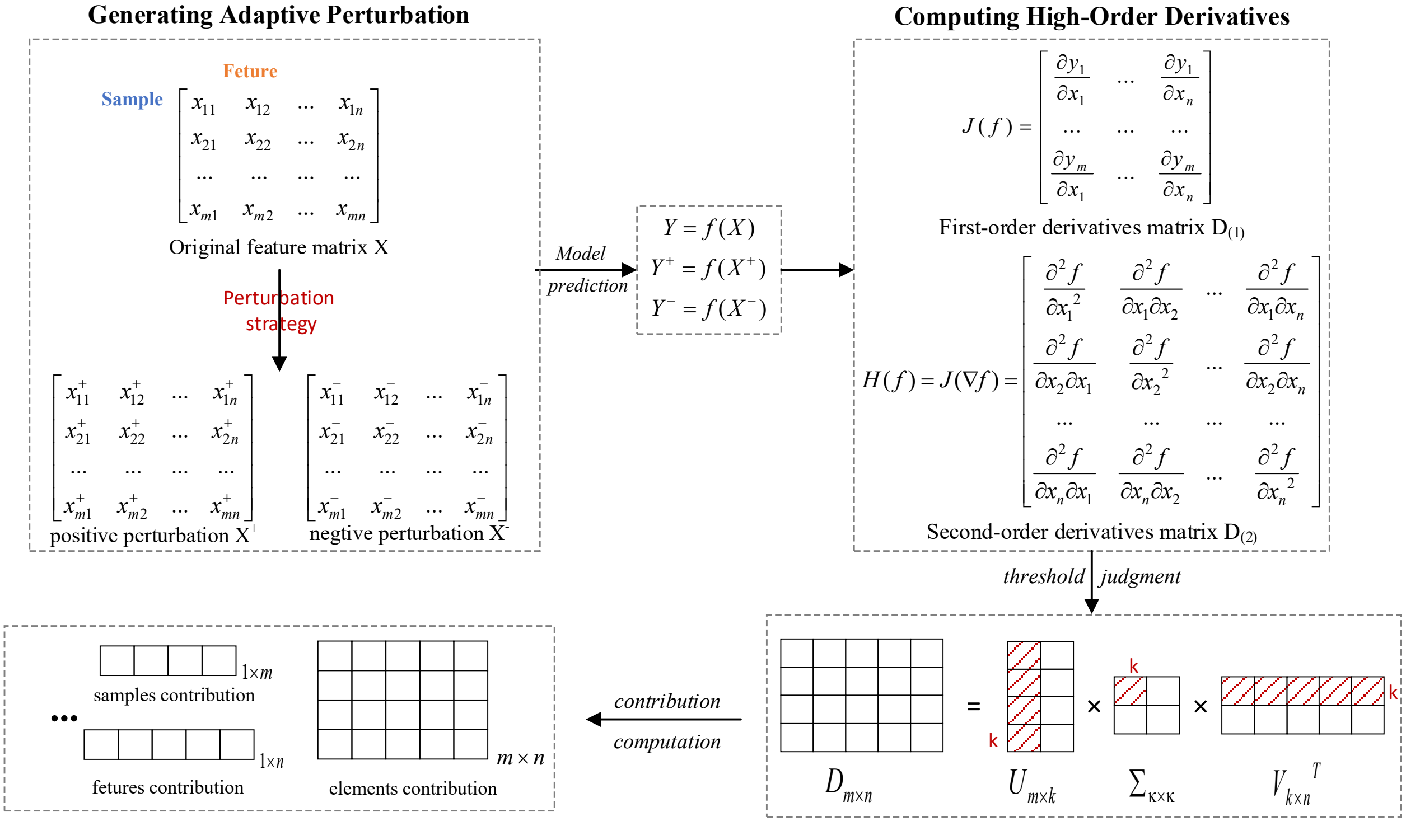

**Fig. 1.** Schematic diagram of the ADORE algorithmic framework.

## 3.1 Generating Adaptive Perturbation

Traditional explainability methods often treat feature importance and instance-specific explanations as separate entities, which may obscure the understanding of how features interact with individual samples to influence predictions. The core idea of ADORE lies in constructing a derivative matrix that captures the sensitivity of model outputs to all input features across samples. This approach enables a unified analysis of both global feature importance and local instance contributions.

Black-box models make direct derivative methods unavailable due to the inability to access their internal computational details or gradients. The derivatives can be approximated by adding or subtracting a small value (i.e., a perturbation) to the input features and observing the output change. Therefore, ADORE needs to provide the adaptive perturbation strategy $\delta$ for different types of input data.

### 3.1.1 Feature extraction and perturbation strategy for tabular data

When the input to the ADORE algorithm is a tabular dataset, it consists of m samples, each with n features. No additional feature extraction step is required. At the perturbation step, ADORE computes the perturbation using two strategies:

$$\delta_j = \varepsilon \times \sigma_j \tag{1}$$

$$\delta_j = \varepsilon \times r_j \tag{2}$$

Where $\delta_j$ denotes the perturbation value of the feature; $\varepsilon$ is a perturbation factor, which is an empirical value; $\sigma_j$ denotes the standard deviation for each feature; $r_j$ indicates the range of features.

It is worth noting that perturbations based on standard deviation are better suited for datasets with Gaussian distributions, as skewed distributions may result in asymmetric perturbations. Conversely, range-based perturbations can be more effective for non-normal features but may become excessively large in the presence of outliers, leading to unrealistic perturbations. In ADORE's design, users are required to select between these approaches based on the characteristics of their dataset.

Then, perturb the $j-th$ feature of the sample will get a positive or negative perturbation:

$$x_{i+j} = x_{i+j} + \delta_j \quad (3)$$

$$x_{i-j} = x_{i-j} - \delta_j \quad (4)$$

### 3.1.2 Feature extraction and perturbation strategy for text data

When the input data is textual, each sample corresponds to a document in the corpus. The ADORE model constructs the feature matrix using a frequency-based Bag-of-Words (BoW) representation, wherein textual semantics are captured by counting the occurrences of individual terms within each document. The CountVectorizer is applied to transform the vocabulary into vector form after removing stop words and low-quality noise terms. Subsequently, the top N most frequent words are retained as features. In the resulting matrix, each entry represents the term frequency for a specific document, thereby indicating the contribution of that word to the document's semantic profile. The upper limit on the number of features, along with other hyperparameters, can be tuned to ensure selection of only the most informative and consistently distributed terms.

During the perturbation phase, ADORE employs semantic expansion to produce both positive and negative perturbations of the identified keywords. Perturbation terms include synonyms and antonyms of the feature words, simulating semantic shifts in the positive and negative directions, respectively. The magnitude of these shifts can be approximated by computing the cosine distance between the embedding vectors of the original and perturbed terms, thus quantifying the semantic displacement within the embedding space.

It should be noted that ADORE is designed for interpreting various black-box models. The input format must therefore be compatible with the target model's requirements. For instance, when applying models such as Support Vector Machines (SVMs) to text classification tasks, the input typically takes the form of a sparse matrix. In such cases, the procedures for feature matrix construction and perturbation follow the same principles as those employed for tabular data.

### 3.1.3 Feature extraction and perturbation strategy for image data

When the input data type is image, ADORE induces perturbations by extracting features from the intermediate hidden layers of a convolutional neural network. In contrast to methods that perturb individual pixels or regions[8], ADORE has the potential to automatically obtain more interpretable and consistent features, such as edges, shapes, and textures. To perform above operations, ADORE first employ the VGG16 model for image feature extraction. Features are extracted using a pre-trained VGG16 model truncated at the pooling layer of the first block. The extracted features consist of multi-channel feature maps, which are reshaped into an individual feature matrix, facilitating further manipulation and analysis.

To study the impact of feature variation, perturbations are applied to the extracted feature matrix. This step involves introducing both positive and negative perturbations to individual feature channels. For each feature column, the perturbation intensity is determined by calculating the average standard deviation across all images, scaled by a small coefficient ε. Perturbations are uniformly applied to all images, resulting in two perturbed feature sets: one with increased values (positive perturbation) and another with decreased values (negative perturbation). This controlled manipulation allows for a systematic investigation of feature sensitivity

Finally, the perturbed feature matrix is used to reconstruct the original images, restoring them to their original dimensions. This is achieved through a series of deconvolution and up-sampling operations. A deconvolution model (Zeiler & Fergus, 2014) is designed to process the feature matrix, using transpose convolutional layers to expand the spatial dimensions and refine the reconstruction. The reconstructed images are converted to RGB format for prediction and visualization. To ensure fidelity, the images are resized back to their original dimensions using the stored size information.

## 3.2 Computing Higher-Order Derivatives

Feature-based perturbations can reveal information about the decision boundaries of the model, however relying on linear approximations may be insufficient to fully capture the model behavior when dealing with complex nonlinear relationships. At this stage, the introduction of higher-order derivative calculations, especially Jacobi and Hessian matrices, emerge as crucial tools for analyzing internal interactions and capturing nonlinear effects within the model.

For models or regions where linear approximation is adequate, the first-order derivative suffices. The Jacobian matrix $J \in \mathbb{R}^{m \times n}$ is computed as:

$$J_{ij} = \frac{\partial f\left(x^{(i)}\right)}{\partial x_j} \tag{5}$$

For capturing nonlinear interactions, the Hessian matrix $H \in \mathbb{R}^{n \times n}$ for each sample $i$ is considered:

$$H_{jk}^{(i)} = \frac{\partial^2 f(x^{(i)})}{\partial x_j \, \partial x_k} \tag{6}$$

where $H^{(i)}$: This denotes the Hessian matrix for sample $i$, showing that it is specific to each input $x^{(i)}$; $j$ and $k$ represent the dimensions of the input features.

### 3.2.1 High-Order approximation

On the basis of considering the perturbation method of eigenvalues, the central difference method is used to quantify the change of eigenvalues before and after the perturbation. Therefore, For each sample $i$ from 1 to $m$, and each feature $j$ from 1to $n$:

Compute the perturbed prediction:

$$y_{i+} = f(x_{i+}) \tag{7}$$

$$y_{i-} = f(x_{i-}) \tag{8}$$

Then the approximated first-order derivatives (Jacobian matrix) can be estimated as:

$$D_{ij} = \frac{y_{i+} - y_{i-}}{2\delta_j} \tag{9}$$

And the approximated second-order derivatives (Hessian matrix) can be estimated as:

$$D_{ij} = \frac{y_{i+} - 2\hat{y}_i + y_{i-}}{\delta_j{}^2} \tag{10}$$

Where $y_{i+}$and $y_{i-}$ are predictions for positive and negative perturbations on the *j-th* feature, respectively.

### 3.2.2 Adaptive higher-order derivatives

To accurately capture both linear and nonlinear feature interactions, ADORE employs an adaptive mechanism that dynamically selects between first-order and second-order derivatives based on the model's local complexity.

(1) Compute a complexity measure: For each sample $x^{(i)}$, compute a complexity measure $C^{(i)}$, such as the magnitude of the second derivative:

$$C(i) = \frac{\left\|\nabla^2 f(x(i))\right\|_F}{1 + \left\|\nabla f(x(i))\right\|_F} \tag{11}$$

where $\|\cdot\|_F$ denotes the Frobenius norm.

(2) Set a Threshold τ: Define a threshold τ to distinguish between linear and nonlinear regions.

(3) Derivative Order Selection: If $C^{(i)} < \tau$,, indicating significant nonlinearity, use the second-order derivative; otherwise, use the first-order derivative.

This adaptive approach ensures accurate interpretation across different regions of the model by capturing the appropriate level of feature interactions.

### 3.3 Implementing Randomized Singular Value Decomposition

Computing and storing the derivative matrices, particularly second-order derivatives, can be computationally expensive for large datasets. ADORE addresses this challenge by employing Randomized Singular Value Decomposition (Randomized SVD)[12], which efficiently approximates the derivative matrix while preserving essential information.

Project the derivative matrix $D$ onto a lower-dimensional subspace using a random matrix $\Omega \in \mathbb{R}^{n\times k}$:

$$Y = D\Omega \tag{12}$$

Compute the SVD of the smaller matrix $Y$:

$$Y = U\Sigma V^T \tag{13}$$

where $U\in\mathbb{R}^{m\times k}$, $\Sigma\in\mathbb{R}^{k\times k}$ and $V\in\mathbb{R}^{k\times k}$.

Approximate the original derivative matrix $D$ using:

$$D \approx U\Sigma V^T \Omega^T \tag{14}$$

The approximation error is bounded and acceptable for explainability purposes, allowing ADORE to handle large-scale data efficiently.

### 3.4 Distinguishing Multidimensional Contributions

The identification of samples and features exerting disproportionate influence on model predictions is crucial for uncovering model vulnerabilities and highlighting areas for improvement. ADORE employs the derivative matrix in conjunction with Randomized Singular Value Decomposition (SVD) to systematically detect these influential elements, thereby enhancing the explainability and robustness of the model.

Understanding not just which features are important but how they influence predictions (positively or negatively) is critical. ADORE quantifies both the direction and magnitude of feature contributions.

For each sample $i$ and feature $j$, the contribution $C_{ij}$ is calculated as:

$$C_{ij} = D_{ij} \tag{15}$$

where $\mu_j$ is the mean of feature $j$ across all samples. The sign of $C_{ij}$ indicates the direction (positive or negative influence), and the magnitude reflects the strength of the contribution.

To assess the relative importance of features within a sample, contributions can be normalized:

$$R_{ij} = \frac{|C_{ij}|}{\sum_{j=1}^{n}|C_{ij}|} \tag{16}$$

This normalization facilitates the comparison of feature contributions across different samples.

### 3.4.1 Leverage scores for samples

To quantify the influence of individual samples, ADORE computes leverage scores $\ell^{(i)}$ based on the SVD-derived components. Specifically, the leverage score for sample $i$ is given by:

$$\ell^{(i)} = \sum_{i=1}^{m} U_{i,j} \tag{17}$$

where $U_{i,:}$ denotes the $i-th$ row of matrix $U$, which contains sample-specific information encapsulated in the left singular vectors. High leverage scores correspond to samples that exert significant impact on model predictions, thus identifying data points critical to the model's behavior.

### 3.4.2 Feature importance scores

Similarly, compute feature importance scores $s_j$:

$$s_j = \sum_{i=1}^{k} V_{i,j} \tag{18}$$

where $V_{i,:}$ is the $i-th$ row of the matrix $\boldsymbol{V}$, capturing feature-specific information through the right singular vectors of the SVD decomposition. Features with high importance scores are identified as having a substantial effect on the model's decisions, marking them as key contributors to the model's predictive behavior.

In summary, leverage scores $\ell^{(i)}$ facilitate the identification of influential samples, while feature importance scores $s_j$ highlight critical features. This dual approach enables ADORE to provide a nuanced understanding of the elements that drive model predictions, supporting a more transparent and interpretable model framework.

Algorithm 1 provides the pseudocode for the ADORE algorithm.

**Algorithm 1: ADORE Algorithm**

| | |
|---|---|
| | **Input:** Trained model $f$, dataset $X = x^{(1)}, x^{(2)}, \dots, x^{(m)}$, threshold $\tau$, sparsity threshold $\varepsilon$ |
| | **Output:** Derivative matrix $D$, contributions $C$, key samples and features |
| **1** | Initialize $D$ as an empty matrix |
| **2** | **for** each sample $x^{(i)}$ in $X$ **do** |
| **3** | Compute first-order derivatives (Jacobian) $J^{(i)}$ using Eq. (6) |
| **4** | Compute second-order derivatives (Hessian) $H^{(i)}$ using Eq. (7) |
| **5** | Compute complexity measure $C^{(i)}$ using Eq. (12) |
| **6** | **if** $C^{(i)} < \tau$ **then** |
| **7** | $D_i = J^{(i)}$ |
| **8** | **else** |
| **10** | $D_i = Flatten(H^{(i)})$ |
| **11** | **end** |
| **12** | Apply sparsity threshold: $D_i[j] = 0 \; if |D_i[j]| < \varepsilon$ |
| **13** | **end** |
| **14** | Stack $D_i$ to form the derivative matrix $D$ |
| **15** | Perform Randomized SVD on $D$: $D \approx U\Sigma V^T$ |
| **16** | Compute contributions $C$ using $D$ and feature means $\mu$ |
| **17** | Identify key samples using leverage scores $\ell^{(i)}$ |
| **18** | Identify key features using importance scores$s_j$ |
| **19** | Return $D$, $C$, key samples, and key features |

## 4. Experimental Setup

This section mainly introduces information related to the experiment, such as task design, model selection, baseline interpretation methods, evaluation metrics, and parameter settings.

### 4.1 Tasks and models

To comprehensively evaluate the explainability performance of ADORE in multimodal and multitask scenarios, this study considers three representative task types—regression, binary classification, and multiclass classification—covering three distinct data modalities: tabular, textual, and image data. The overall experimental scope is summarized in Table 1.

For the tabular modality, we adopt the California Housing dataset, comprising 20,640 records with eight key features. A Random Forest regression model is employed, with an 8:2 train–test split. During training, critical hyperparameters are tuned via grid search (GridSearchCV), including the number of decision trees (n_estimators = 300), maximum tree depth (max_depth = 30), and minimum samples required for node splitting (min_samples_split = 5). The final model achieves a root mean squared error (RMSE) of 0.49 and a coefficient of determination ($R^2$) of 0.82, and is subsequently selected as the analysis target for the house price prediction task.

For the text modality, we utilize the IMDB Reviews dataset, which contains 50,000 movie reviews (balanced between positive and negative sentiment). The task is binary sentiment polarity classification. The target model is the bert-base-uncased-finetuned-imdb checkpoint provided by Hugging Face, which attains an accuracy of 91% on the official test split and can be directly used in subsequent explainability experiments.

For the image modality, we use the CIFAR-10 dataset, comprising 60,000 RGB images of size 32×32 across 10 everyday object categories. The model of choice is ResNet-50 (pretrained on ImageNet and subsequently fine-tuned on CIFAR-10). The training strategy adopts SGD with momentum (learning rate = 0.1, epochs = 100, batch size = 128), yielding a test accuracy of 94.7%. This ensures that the deep convolutional features learned by the CNN are suitable for subsequent explainability evaluation.

**Table 1.** Overview of experimental tasks.

| Modality | Task | Dataset | Model Type | Task description |
|---|---|---|---|---|
| Tabular | Housing price prediction | California Housing | Random Forest regression | Predict the median value (in $1000) of owner-occupied homes from 8 demography and housing attributes. |
| Text | Sentiment classification | IMDB reviews | fine-tuned BERT | Determine the binary sentiment polarity (positive/negative) of 50k movie reviews. |
| Image | Object recognition | CIFAR-10 | ResNet50 | Identify the object category of a given image from 10 possible classes in the CIFAR-10 dataset. |

### 4.2 Baseline Methods and Evaluation Metrics

To establish robust baselines for interpretability analysis, we selected a set of widely adopted post-hoc explanation techniques, covering both local and global perspectives. These methods have been extensively applied across diverse machine learning tasks and are known to provide interpretable feature importance scores for regression models. Specific methods and settings can be found in Table 1.

**Table 1.** Baseline methods and their specific settings.

| Baseline method | Settings |
|---|---|
| LIME | For each instance, we construct a local linear surrogate with kernel width set to $0.75\sqrt{d}$ (d denotes feature dimensionality). The requisite perturbations are generated 5 000 times for tabular data and 3 000 times for text or image modalities. |
| SHAP | A background data set of k = 100 samples is employed; convergence is declared after $2^{10}$ weighted subset evaluations, guaranteeing stable Shapley value estimates. |
| PFI | Every input feature is independently shuffled while keeping the remaining covariates intact. The resultant deterioration in predictive performance—quantified on the validation set via $R^2$ and RMSE—is averaged across multiple random permutations to yield an importance score. |
| IG | Feature attributions are computed via path integration of model gradients from a baseline vector of all zeros to the original input. The integration path is approximated using 50 equally spaced interpolation steps; gradients are averaged across these steps to obtain stable estimates. |

To enable a rigorous comparison of explanation methods across different data modalities and task types, we designed a set of quantitative metrics. Unlike model accuracy, the performance of explainability methods cannot be directly measured; thus, we define evaluation criteria along five complementary dimensions: fidelity, stability, sparsity, and computational efficiency. The specific design and computation for each dimension are as follows.

- Fidelity (FID): Fidelity quantifies the extent to which the identified important features are consistent with the model's decision-making process. For each explanation method, we mask (either remove or replace with the mean value) the top-k most important features determined by that method, and record the percentage drop in model performance. $R^2$ is used for regression tasks, binary-F1 score for sentiment classification, and accuracy for image recognition tasks.
- Stability (STB): Stability measures the sensitivity of explanations to small perturbations or randomness. For each test sample x, a perturbed version is generated by adding noise $\epsilon \sim \mathcal{N}(0, \sigma^2)$, and the Spearman correlation coefficient between the two sets of explanation results is computed. A value closer to 1 indicates higher stability. For textual data, perturbations involve replacing 5–15% of tokens with stopwords; for image data, pixel-level Gaussian noise is applied.

- Sparsity (SPR): Sparsity reflects the compressibility of explanations—whether a few key features can sufficiently capture the decision logic. We compute the cumulative contribution rate of the top-k features by ranking feature importance for each sample and calculating the proportion of overall importance accounted for by the top-k. If a small number of features account for a large proportion of importance, the explanation is considered sparse and highly interpretable.
- Computational Efficiency (Eff): Efficiency is assessed by measuring the wall-clock runtime and peak memory usage during explanation generation on a single CPU core (Intel Xeon Gold 6248 @ 2.5 GHz).

### 4.3 Parameter and Computational Environment Settings

The ADORE algorithm contains several important parameters. Table 2 gives the parameter configurations for the experiment along with the default value ranges.

**Table 2.** ADORE parameter configurations.

| No. | Parameter | Range/Value | Description |
|---|---|---|---|
| 1 | Derivative Order Threshold ($\tau$) | $[0, +\infty)$ | Determines the switch between first-order and second-order derivatives. It was set based on preliminary experiments to balance accuracy and computational cos. |
| 2 | Sparsity Threshold ($\epsilon$) | 0.001 | A small value to eliminate negligible derivative values, enhancing sparsity in the derivative matrix and improving computational efficiency. |
| 3 | Approximation Rank (k) | $[1, \min(m, n)]$ | A value to balance approximation accuracy. Chosen to retain at least 95% of the cumulative variance, determined via the cumulative explained variance plot. |
| 4 | Number of Oversamples | 10 | Set to enhance the accuracy of the approximation, as recommended in randomized SVD literature[12]. |

The experiment was conducted on a workstation equipped with an Intel Xeon processor and 512 GB of RAM, providing the necessary computational power for training complex models and efficiently performing derivative computations. On the software side, the experiment was implemented using Python 3.8, with libraries such as NumPy, SciPy, and scikit-learn for handling traditional machine learning models, and TensorFlow for neural network models. Sparse matrix operations and Randomized SVD were facilitated by optimized routines available in these libraries, ensuring computational efficiency and accuracy.

## 5 Experimental Results

### 5.1 Tabular Data: House Price Prediction

In the housing price prediction task, four interpretation methods were employed to analyze the test set, assessing the importance of eight features with respect to the model output (Fig. 1). LIME and SHAP provides "local feature contributions" for individual samples; therefore, we applied LIME and SHAP to multiple instances and aggregated the of the contribution values, using this aggregated measure as a proxy for global feature importance.

As illustrated in Fig. 1, in contrast to PFI, which can only quantify the magnitude of feature importance, LIME, SHAP, and ADORE are able to capture the directionality of each feature's influence on the prediction. Nevertheless, these three methods exhibit notable discrepancies in the inferred contribution directions for certain features. Here, "contribution direction" refers to whether an increase in a feature's value typically leads to an increase or decrease in the predicted house price (the target variable). From a domain knowledge perspective, *MedInc* and *AveRooms* are expected to have a strong positive effect, whereas *HouseAge*, *AveBedrms*, *Population*, and *AveOccup* are expected to have negative effects. Our results indicate that ADORE produces contribution directions most consistent with such expectations.

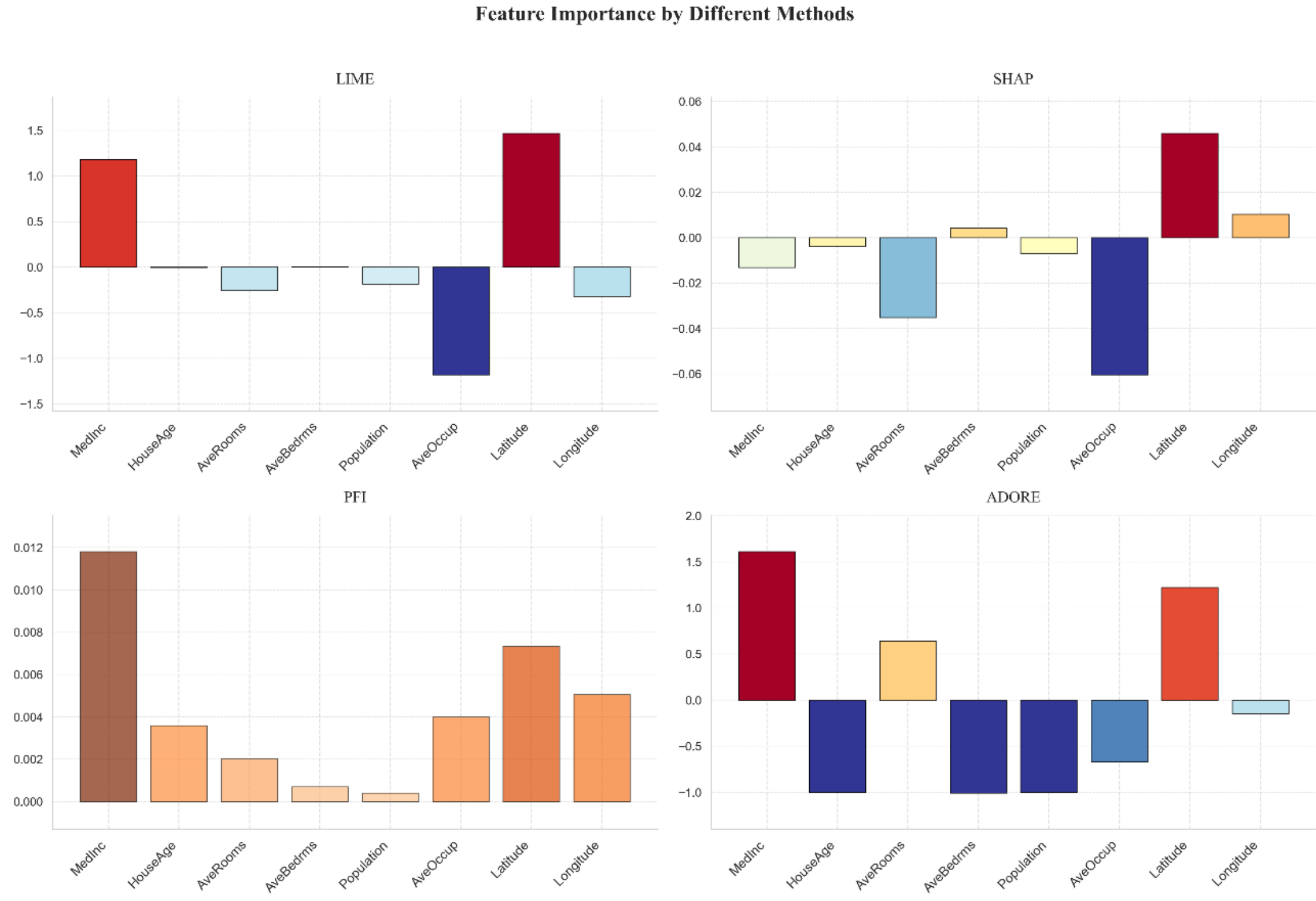

**Fig. 2.** Comparison of feature importances by different methods.

Table 3 summarizes the performance metrics of all interpretation methods. In terms of fidelity (k = 2), ADORE and PFI both achieve an FID score of 98.91%, substantially higher than LIME and SHAP (both 57.26%). This identical result stems from different methods identifying the same top-2 influential features, indicating stronger consistency between the explanations and actual feature contributions for ADORE and PFI, whereas LIME and SHAP may introduce bias in feature importance estimation. Regarding stability, PFI attains the highest STB score (71.43%), reflecting highly consistent explanation behavior. ADORE achieves 53.18%, markedly outperforming LIME and SHAP, suggesting lower sensitivity of its explanations to minor input perturbations. In sparsity, assessed via the top-k cumulative contribution rate, all methods show comparable performance, demonstrating that a small subset of key features can account for nearly all predictive contributions. In computational efficiency, SHAP and PFI incur higher time and memory costs, whereas ADORE maintains low computational complexity. Notably, ADORE achieves a favorable trade-off among interpretability accuracy (FID), stability (STB), sparsity (SPR), and efficiency, highlighting its robust and well-balanced performance across diverse evaluation criteria.

**Table 3.** All performance indicators of the interpretation methods in the house price prediction task.

| Methods | FID (k=2) % | STB % | SPR (k=2) % | SPR (k=5) % | Avg Time (s) | Peak Mem (MB) |
|---|---|---|---|---|---|---|
| LIME | 57.26 | 48.42 | 58.43 | 98.08 | **5.83** | 2.81 |
| SHAP | 57.26 | 8.34 | 50.11 | 92.42 | 19.03 | 3.74 |
| PFI | **98.91** | **71.43** | **61.72** | 96.18 | 23.36 | 367.88 |
| ADORE | **98.91** | 53.18 | 52.67 | **99.05** | 9.80 | **0.39** |

## 5.2 Text Data: sentiment classification

PFI is applicable to text classification tasks only when the features are numerically permutable (e.g., TF–IDF vectors). For deep representations such as raw inputs to Transformers or BERT, the notion of "permutation" is ill-defined; therefore, in this section we adopt the IG method, which is well-suited for explaining neural networks. To demonstrate the performance differences among various explanation techniques in the context of sentiment classification for movie reviews, we selected a review containing a pronounced rhetorical shift, which the prediction model classified as positive. Figure 3 presents the visualizations of four explanation methods applied to this example.

For LIME, SHAP, and ADORE, visualizations were generated using their respective built-in interpretation tools, where individual tokens are distinguished by different colors for positive and negative contributions, and the color intensity corresponds to the magnitude of the attribution score. In contrast, the IG visualization was manually rendered based on computed attribution values, following the same visual style as the other methods to facilitate consistent comparison.

As illustrated, with the exception of LIME—which yields localized and sparse attribution results—the other methods successfully identified key phrases such as brilliant, gripping, and loving as major contributors to the positive prediction, while marking terms such as slow-paced, confusing, and unnecessary as negative contributors. Notably, ADORE and SHAP achieve broader contextual coverage, yet differ in their sensitivity to feature dependencies. ADORE reduces spurious attributions to function words or neutral terms, exhibiting superior interpretability and alignment with human intuition. In contrast, IG distributes scores more diffusely across multiple tokens, introducing more pronounced attribution noise.

**LIME**

The first 30 minutes were kinda slow-paced and confusing—some scenes felt unnecessary. But the second half hit hard: the acting was truly brilliant, the plot twists were gripping, and the ending left me deeply emotional. Ended up loving this movie!

**SHAP**

The first 30 minutes were kinda slow-paced and confusing—some scenes felt unnecessary. But the second half hit hard: the acting was truly brilliant, the plot twists were gripping, and the ending left me deeply emotional. Ended up loving this movie!

**IG**

The first 30 minutes were kinda slow-paced and confusing—some scenes felt unnecessary. But the second half hit hard: the acting was truly brilliant, the plot twists were gripping, and the ending left me deeply emotional. Ended up loving this movie!

**ADORE**

The first 30 minutes were kinda slow - paced and confusing —some scenes felt unnecessary . But the second half hit hard : the acting was truly brilliant , the plot twists were gripping , and the ending left me deeply emotional . Ended up loving this movie !

**Fig. 3.** Explanation visualizations on a sentiment classification example.

Similarly, we evaluated the performance metrics of different methods on the sentiment classification task. ADORE performs excellently in fidelity, stability, and sparsity, indicating that its explanations are highly focused on key semantic units and exert stronger causal influence and robustness on the model's decisions. In contrast, IG produces dispersed attributions with low fidelity; LIME shows limited stability and faithfulness; while SHAP achieves performance close to ADORE, it incurs higher computational overhead. The results demonstrate that ADORE delivers reliable, concise, and human-intuitive explanations while maintaining high efficiency.

**Table 4.** All performance indicators of the interpretation methods in the sentiment classification task.

| Methods | FID (k=10) % | STB % | SPR (k=5) % | SPR (k=10) % | Avg Time (s) | Peak Mem (MB) |
|---|---|---|---|---|---|---|
| LIME | 20.00 | 64.33 | 63.00 | 87.46 | 300.51 | **145.46** |
| SHAP | 28.00 | **78.30** | 75.55 | 90.54 | 91.61 | 333.19 |
| IG | 25.00 | 70.71 | 68.52 | 85.72 | 65.00 | 200.41 |
| ADORE | **29.00** | 73.45 | **88.30** | **97.62** | **14.09** | 218.34 |

### 5.3 Image Data: Object Recognition

For the object recognition task, we take a panda image from the dataset as an example. Fig. 4 illustrates the results of interpretability analysis for the classification decision of the ResNet50 model using three different methods. LIME generates importance masks based on superpixel perturbations, spatially highlighting several local regions attended to by the model. However, some of these highlighted regions appear as relatively coarse block-wise patterns, lacking a direct correspondence to high-level semantic features. SHAP constructs pixel-level heatmaps based on feature contribution values, quantitatively characterizing the positive and negative contributions of each position to the prediction outcome. Nevertheless, this visualization suffers from limitations in spatial resolution and semantic readability, making it difficult to reveal the specific meaning of deep features inside the model. ADORE focuses on selecting the top-five feature maps from the model's high-level feature representations that have the greatest influence on the prediction, and visualizes their spatial saliency distributions. This enables the method to clearly depict core patterns of model attention at the semantic level — including edges (Feature 2 and Feature 52), texture patterns (Feature 29), salient object parts (Feature 18 and Feature 27), and background information (Feature 18). The approach not only preserves the high semantic resolution inherent to deep features, but also enhances the intuitiveness and interpretive value of the results, thereby revealing the mechanism by which the model leverages multi-level information for decision-making in classification tasks.

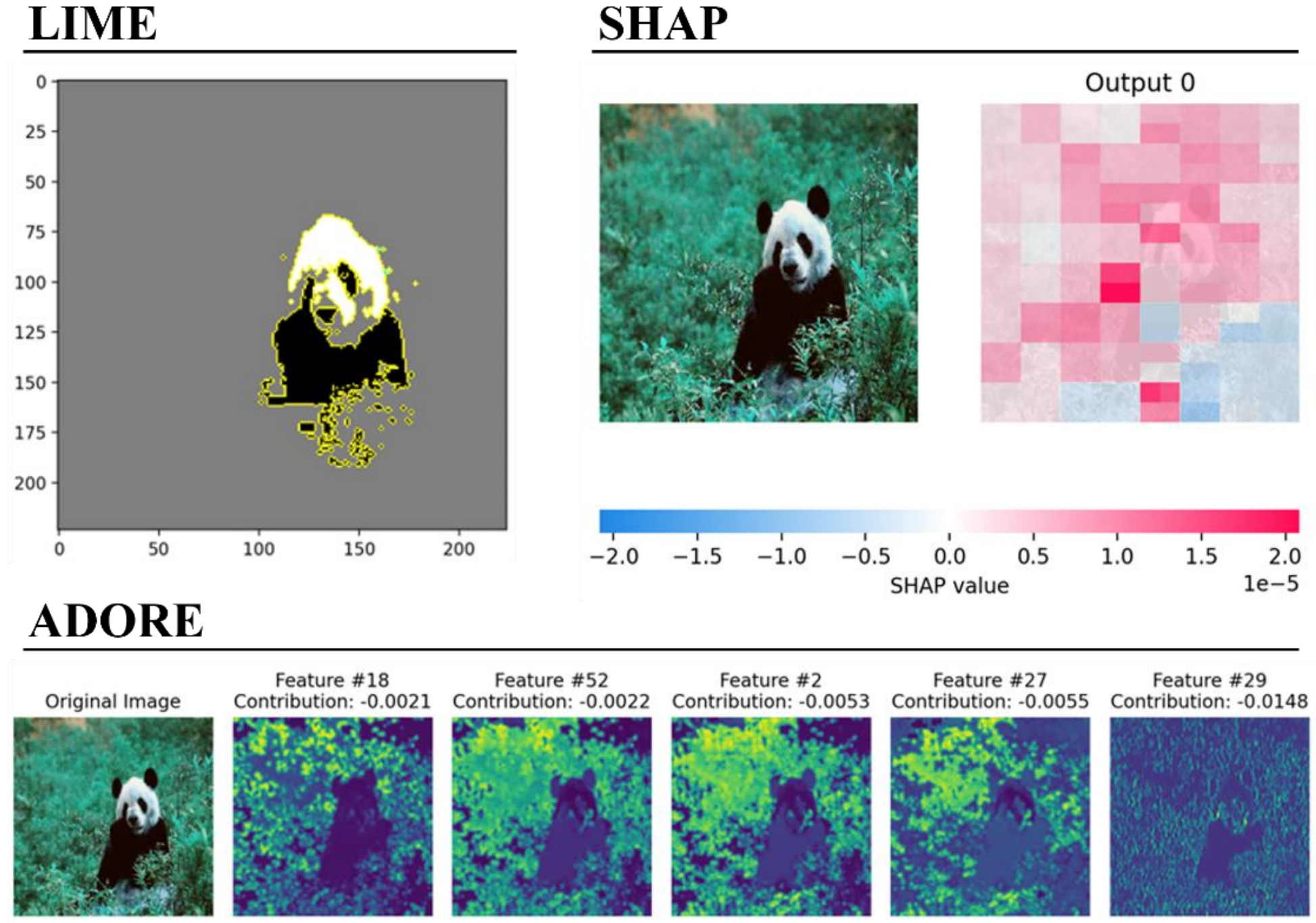

**Fig. 4.** Comparative visual explanations of object recognition results using LIME, SHAP, and ADORE.

Further, ADORE algorithm demonstrates insensitivity to variations in model architecture, yielding approximate feature contribution results across different model's settings. As showed in Fig.5, the upper result shows the interpretation results of the VGG16 model with default parameters on ImageNet using the ADORE method, while the bottom shows the explanation results for the VGG16 model trained on CIFAR-10 data. The top 5 feature contributions identified between the two are the same. This characteristic indicates that ADORE provides high stability in its interpretations, offering users reliable information on feature contributions without significant alterations in the interpretation results due to model adjustments. This stability is particularly valuable in practical applications, as it ensures the reliability of model interpretation. Regardless of different model configurations, ADORE can consistently provide coherent interpretations.

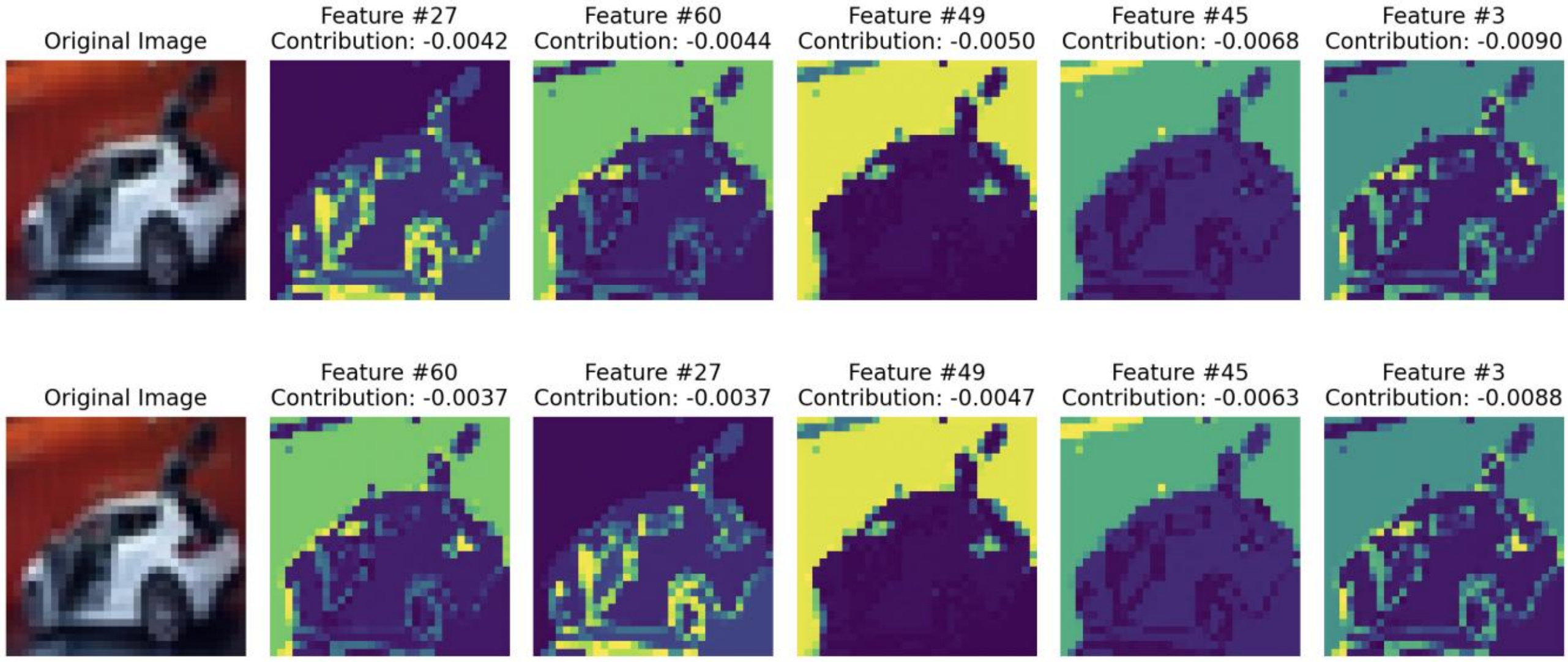

**Fig. 5.** Top 5 feature contributions of images obtained by different prediction model weights

## 6. Discussion

### 6.1 Computational Complexity Analysis

In this section, we present a comprehensive computational complexity analysis of the ADORE algorithm. This analysis incorporates the use of approximate methods for computing Hessian and Jacobian matrices and considers the prediction model's complexity, denoted as $C_f$. Our goal is to provide a clear understanding of the algorithm's efficiency and identify the primary factors that influence its computational cost. For each of the $m$ samples in the dataset, the algorithm performs several computations.

#### 6.1.1 Complexity measure computation

For each of sample $x^{(i)}$, the algorithm computes the complexity measure $C^{(i)} = \left\|\nabla^2 f\left(x^{(i)}\right)\right\|_F$ using the approximated Hessian matrix. This process involves: predicting the label at the original sample to obtain $\hat{y}_i = f(x^{(i)})$; perturbing each of the $n$ features positively and negatively, resulting in $2n$ additional samples; predicting the label at each perturbed sample to obtain $y_{i+}$ and $y_{i-}$ for all features.

The total number of model evaluations per sample is therefore $1 + 2n$. Then the total computational complexity for computing the complexity measure per sample is $O(n \cdot C_f)$.

#### 6.1.2 Threshold comparison and derivative computation

Threshold Comparison and Derivative Computation The algorithm then compares $C^{(i)}$ with the threshold $\tau$ The algorithm computes the approximated second-order derivatives (Hessian matrix) and the approximated first-order derivatives (Jacobian matrix) using the previously obtained $\hat{y}_i$, $y_{i+}$ and $y_{i-}$. In both cases, the derivative computation does not involve additional model evaluations and thus does not add to the $O(n \cdot C_f)$ complexity established during the complexity measure computation.

### 6.1.3 Sparsity thresholding

After computing the derivatives, the algorithm applies the sparsity threshold $\varepsilon$ to the derivative vector $D_i$. This involves iterating over all $n$ elements of $D_i$ and zeroing out those below the threshold. The complexity of this operation is $O(n)$ per sample.

### 6.1.4 Summary of sample processing

Combining the complexities of the steps above, the total computational complexity per sample is dominated by the model evaluations, amounting to $O(n \cdot C_f)$. Therefore, for all $m$ samples, the cumulative complexity is $O(m \cdot n \cdot C_f)$.

### 6.1.5 Formation of the derivative matrix

Once all samples have been processed, the algorithm stacks the derivative vectors $D_i$ to form the derivative matrix $D$ of size $m \times n$. The stacking operation involves organizing the data into a matrix structure and has a complexity of $O(m \cdot n)$, which is relatively small compared to the previous steps.

### 6.1.6 SVD

The algorithm performs a randomized SVD on the derivative matrix $D$ to reduce its dimensionality and extract significant patterns. The computational complexity of this step is $O(m \cdot n \cdot k)$ , where $k$ is the target rank for the SVD and typically $k \ll m$ and $k \ll n$.

### 6.1.7 Contributions computation

Using the results from the SVD, the algorithm computes the element contributions $C$ by combining the derivative matrix $D$ with the feature means $\mu$.This step has a complexity of $O(m \cdot n)$, as it involves matrix-vector multiplications and element-wise operations over $D$.

Finally, the algorithm identifies key samples and features:

(i) Key Samples: Calculated using leverage scores derived from the SVD components. The complexity of computing these scores is $O(m \cdot k)$.

(ii) Key Features: Determined using importance scores, which require $O(n \cdot k)$ computations.

These steps involve relatively simple operations compared to earlier stages and do not significantly impact the overall complexity.

### 6.1.8 Total computational complexity

The total complexity is the sum of the complexities of each step above because they are executed sequentially. We only need to focus on the main complexity terms, namely $O(m \cdot n \cdot C_f)$ and $O(m \cdot n \cdot k)$, and add them to obtain the total complexity. The complexities of other steps are negligible compared to the two main complexities. Therefore, the total computational complexity of the ADORE algorithm is $O(m \cdot n \cdot C_f + m \cdot n \cdot k)$. The computational complexity analysis reveals that the ADORE algorithm's efficiency is primarily influenced by the number of samples $m$, the number of features $n$, the complexity of the model's prediction $C_f$, and the target rank $k$ for the randomized SVD. The dominant computational cost arises from the model predictions required for approximating the derivatives using finite differences.

## 6.2 Theoretical and Practical Significance

From a theoretical perspective, the introduction of ADORE enriches the methodologies for interpretable machine learning. Existing interpretability approaches tend to focus on separately analyzing global feature importance and local sample contributions. ADORE provides both global and local explanations within a cohesive framework. By analyzing derivative matrices, it captures both global feature importance and local sample contributions, offering a novel theoretical framework for comprehensive model analysis. Through quantifying complex relationships between features and samples, ADORE provides theoretical tools to analyze sources of bias and vulnerabilities in models. This capability not only supports model optimization but also underpins research in fairness and bias control. Moreover, the adaptability and generalizability of the ADORE framework establish a robust theoretical foundation for interpretability research across different data modalities and model types.

From a practical perspective, ADORE enhances model transparency and users' trust in black-box model decisions by providing precise and comprehensive explanations of machine learning models. Practitioners can leverage ADORE's insights to identify model weaknesses, understand feature contributions, and improve models, thereby enhancing performance and reliability. Furthermore, its efficiency and adaptability enable seamless integration into industry workflows, promoting the practical application of interpretable AI. Finally, we have made the ADORE method publicly available and developed an integrated Python package to facilitate its use by the research community. This open-source implementation enables other scholars to reproduce and build upon our findings, fostering transparency, reproducibility, and further advancements in the field of interpretability research.

## 6.3 Limitations

While ADORE demonstrates significant advancements, several limitations and challenges warrant attention. A primary concern is scalability to extremely large datasets, as the computational demands of derivative calculations and SVD, even with randomized methods, can pose challenges in big data scenarios, underscoring the need for further optimization. Additionally, the algorithm's computational resource requirements present a limitation, particularly for high-order derivative computations such as second-order derivatives, which can be resource-intensive and may restrict applicability in environments with constrained computational resources, necessitating compromises in derivative order selection. Finally, the reliance on randomized SVD introduces approximation errors in derivative matrix reconstruction, which, while generally acceptable for explainability, may affect the accuracy of explanations in applications demanding high precision.

### 6.4 Future Directions

To address the limitations of ADORE and enhance its capabilities, future research can focus on three key areas. First, algorithmic optimization could be explored by leveraging advanced technologies such as distributed computing frameworks, GPU acceleration, and efficient approximation methods like randomized numerical linear algebra, thereby improving scalability and performance on extremely large datasets. Second, the development of automated parameter tuning methods is critical for dynamically selecting optimal values for parameters such as the derivative order ($\tau$) and sparsity ($\epsilon$), guided by data characteristics or specific explainability requirements; machine learning algorithms could be employed to adaptively adjust these parameters during runtime. Lastly, extending ADORE to multi-modal datasets would enable it to analyze diverse data types, including images, text, audio, and time-series data, significantly broadening its applicability to domains such as multimedia analysis, natural language processing, and financial forecasting. These advancements would collectively enhance the robustness, adaptability, and utility of ADORE in complex analytical scenarios.

## 7. Conclusion

This paper introduced ADORE, a novel algorithm designed to enhance the interpretability of black-box machine learning models. ADORE provides a unified framework that captures both global feature importance and local sample contributions through the construction of a derivative matrix. Its adaptive mechanism for selecting the derivative order enables accurate interpretation of both linear and highly nonlinear models, ensuring that complex feature interactions are effectively captured.

Efficient computation is achieved through the integration of Randomized SVD and dynamic sparsity detection and handling. These methods significantly reduce computational complexity and memory consumption, making ADORE both scalable and practical for large-scale applications. The algorithm's model-agnostic nature, relying solely on input-output evaluations, further expands its applicability across a diverse range of machine learning models, including neural networks, ensemble methods, and support vector machines.

Extensive experiments conducted on various datasets and models demonstrated that ADORE not only outperforms baseline interpretability methods in capturing nonlinear interactions and computational efficiency but also provides deeper insights into model decisions. The algorithm's robustness and reliability were affirmed through stability analyses, showcasing its resilience to data perturbations and model parameter changes.

## Appendices

### Appendix A: Notation and Symbols

$f$: Predictive model function.

$x^{(i)}$:$i-th$ sample in the dataset.

$x_j$: $j-th$ feature.

$D$: Derivative matrix.

$J$: Jacobian matrix.

$H$: Hessian matrix.

$C(i)$:Complexity measure for sample $i$.

$\tau$: Threshold for derivative order switching.

$\epsilon$: Sparsity threshold.

$U, \Sigma, V$: Matrices from Singular Value Decomposition.

$C_{ij}$: Contribution of feature $j$ to sample $i$'s prediction.

$\ell(i)$: Leverage score for sample $i$.

$s_j$: Importance score for feature $j$.

## Acknowledgments

The authors would like to thank the National Natural Science Foundation of China for their support of this research under the project "Data Storytelling Methods and Key Technologies for Predictive Analysis Results" (Project No. 72074214).